\def\year{2021}\relax
\documentclass[letterpaper]{article} 
\usepackage{aaai21}  
\usepackage{times}  
\usepackage{helvet} 
\usepackage{courier}  
\usepackage[hyphens]{url}  
\usepackage{graphicx} 
\usepackage{natbib}  
\usepackage{caption} 

\usepackage{amsmath}        
\usepackage{amsthm}
\usepackage{bm} 
\usepackage{subfigure} 
\usepackage{tabularx} 
\usepackage{float}
\usepackage{mathtools}
\usepackage{enumitem} 
\usepackage{amssymb}
\usepackage{pifont}

\title{A Unified Taylor Framework for Revisiting Attribution Methods}
\author {
    Huiqi Deng\textsuperscript{\rm 1,}\footnote{This work is done during her visit at Texas A\&M University.
},
    Na Zou\textsuperscript{\rm 2},
    Mengnan Du\textsuperscript{\rm 2},
    Weifu Chen\textsuperscript{\rm 1,}\thanks{Corresponding author},
    Guocan Feng\textsuperscript{\rm 1},
    Xia Hu\textsuperscript{\rm 2}
    \\
}
\affiliations {
    \textsuperscript{\rm 1} Sun Yat-Sen University \\
    \textsuperscript{\rm 2} Texas A\&M University

    \{denghq7@mail2,mcsfgc@mail,chenwf26@mail\}.sysu.edu.cn,
    \{nzou1,dumengnan,xiahu\}@tamu.edu
}

\begin{document}
	\newcolumntype{L}[1]{>{\raggedright\arraybackslash}p{#1}}
	\newcolumntype{C}[1]{>{\centering\arraybackslash}p{#1}}
	\newcolumntype{R}[1]{>{\raggedleft\arraybackslash}p{#1}} 
\maketitle

\begin{abstract}
Attribution methods have been developed to understand the decision making process of machine learning models, especially deep neural networks, by assigning importance scores to individual features. Existing attribution methods often built upon empirical intuitions and heuristics. There still lacks a general and theoretical framework that not only can unify these attribution methods, but also theoretically reveal their rationales, fidelity, and limitations. To bridge the gap, in this paper, we propose a Taylor attribution framework and reformulate seven mainstream attribution methods into the framework.
Based on reformulations, we analyze the attribution methods in terms of rationale, fidelity, and limitation.
Moreover, We establish three principles for a good attribution in the Taylor attribution framework, i.e., low approximation error, correct contribution assignment, and unbiased baseline selection.
Finally, we empirically validate the Taylor reformulations, and reveal a positive correlation between the attribution performance and the number of principles followed by the attribution method via benchmarking on real-world datasets.
\end{abstract}

\section{Introduction}
Attribution methods have become an effective computational tool in understanding the behavior of machine learning  models, especially Deep Neural Networks (DNNs)
\cite{du2019techniques,samek2019explainable}. They uncover how machine learning models make a decision by calculating the contribution score of each input feature to the final decision. For example, in image classification, the attribution methods infer the contribution of each pixel to the predicted label for a pre-trained model, and usually create saliency maps to visualize the contributions.

Although several attribution methods \cite{samek2020toward} have been proposed recently,
they are based on different heuristics and have very limited theoretical understanding and support.
For instance, Occlusion-1 and Occlusion-patch observe the changes of the output induced by adjusting each input pixel or patch \cite{zeiler2014visualizing,zintgraf2017visualizing};
Layer-wise Relevance Propagation (LRP) evaluates the contributions of each input to a non-linear neuron according to the corresponding linear weights in the pre-trained model \cite{bach2015pixel}.  The interpretations generated by those attribution methods with such intuitive rationales are difficult to compare and can not be fully trusted. Hence, it’s highly desirable to conduct a comprehensive investigation on the rationales, fidelity, and limitations of those various heuristic methods.
Specifically, the following important questions need theoretical investigation:
 \textbf{Rationale}--\textit{What model behaviors do these attribution methods actually reveal};
 \textbf{Fidelity}--\textit{How much can decision making process be attributed};
 \textbf{Limitations}--\textit{Where they may fail}.

While some attempts have been made to partially answer the questions by unifying a certain kind of attribution methods, such as additive feature attribution \cite{lundberg2017unified}, multiplying a modified gradient with input \cite{Marco2018Towards}, or first-order Taylor expansion \cite{samek2020toward}, the problems are still not addressed well due to two challenges.
The first challenge (\textbf{Ch1}) stems from the fact that it is difficult to unify most of existing attribution methods, i.e., to reformulate these attribution methods into one framework, because the methods are based on various heuristics, as we discussed above.
The second challenge (\textbf{Ch2})
is lacking a theoretical attribution framework, which could offer a good description to the attribution problem so as to theoretically reveal the rationale, fidelity, and limitations of the attribution methods.

In this paper, we study the attribution of DNNs to answer the aforementioned three questions by proposing a general Taylor attribution framework and unifying seven mainstream attribution methods into the framework.
The basic idea behind the proposed framework is to attribute an approximation function of DNNs, instead of DNNs themselves.
The proposed Taylor attribution framework has three features:
(1) the framework is based on Taylor expansion, which is able to approximate sufficiently the behavior of black-box DNNs and has a theoretical guarantee on approximation error;
(2) Taylor expansion is a polynomial function, in which attribution and analysis become very easy and intuitive;
(3) The Taylor attribution framework is very general, it can unify many attribution methods that are to analyze the output change between input sample and baseline.

We then reformulate seven mainstream attribution methods into the proposed Taylor attribution framework by theoretical derivations.
The unified reformulations enable us to examine rationales, measure fidelity, and reveal limitation for the existing attribution methods in a systematic and theoretical way.
We analyze the seven attribution methods by their reformulations.
Based on the reformulations and analysis, we establish and advocate three principles for a good attribution in the Taylor attribution framework, which are low approximation error, correct contribution assignment, and unbiased baseline selection.

Finally, we empirically validate the proposed Taylor reformulations by comparing the attribution results obtained by the original attribution methods and their Taylor reformulations.
The experimental results on MNIST show the two attribution results are almost consistent.
We also reveal a strong positive correlation between the attribution performance and the number of principles followed by the attribution method via benchmarking on MNIST and Imagenet.
In summary, this paper has three main contributions:
\begin{itemize}
\item We propose a general Taylor attribution framework, and theoretically reformulate seven mainstream attribution methods into the  attribution framework.
\item We analyze the attribution methods by their reformulation in terms of rationale, fidelity, and limitation, and accordingly establish three principles for a good attribution.
\item We empirically validate the Taylor reformulations, and reveal the relationship between attribution performance and the three principles on MNIST and Imagenet.
\end{itemize}

\section{A General Taylor Attribution Framework}
In this section, we propose a Taylor attribution framework to understand the decision making process of DNNs. Specifically, given a pre-trained DNN model $f$ and an input sample $\bm{x}  = [x_1,x_2,\dots, x_n]^{\rm T}\in \mathbb{R}^n$, the framework aims to infer the contribution of each feature $x_i$ to the prediction $f(\bm{x})$.
We employ a Taylor expansion function $g$ to approximate the DNN model $f$, and then conduct the attribution in $g$ because $g$ is a polynomial function and easy to attribute.

\begin{table}[t]
\centering
\begin{tabular}{L{1.5cm}|L{6.05cm}}
 \textbf{Symbol} & \textbf{Description}   \\ \hline
$f(\bm{x})$ & DNN model with input $\bm{x}$ \\ \hline
$g_K$ & $K$-order Taylor expansion function of $f$ \\ \hline
$\epsilon(\bm{\tilde x},K)$ & Taylor approximation error of $g_K$ at $\bm{\tilde x}$ \\ \hline
$H_{\bm{x}}$ & Hessian matrix at $\bm{x}$ \\ \hline
$H^d_{\bm{x}}, H^t_{\bm{x}}$ & Hessian independent and interactive matrix \\ \hline
$T^\alpha, T^\beta, T^\gamma$ & Taylor first, second, and high-order terms \\ \hline
$T^{{\beta}_{d}},T^{\gamma_d}$ &  Second, high-order independent terms \\ \hline
$T^{{\beta}_{t}}, T^{{\gamma}_{t}}$ & Second, high-order interactive terms \\ \hline
$a_i$ & Attribution of feature $x_i$ \\ \hline
$a_i^{\alpha}$ & Attribution of feature $x_i$ from $T^\alpha$ \\ \hline
$a_i^{\beta_d}, a_i^{\beta_t}$ &  Attribution of feature $x_i$ from $T^{\beta_d},T^{\beta_t}$  \\ \hline
$a_i^{{\gamma}_d},a_i^{{\gamma}_t}$ & Attribution of feature $x_i$  from $T^{\gamma_d},T^{\gamma_t}$  \\ \hline
\end{tabular}
\caption{Symbol descriptions in this paper.}
\label{tab:my_label}
\end{table}

\begin{figure*}[t]\centering
\includegraphics[scale=0.3]{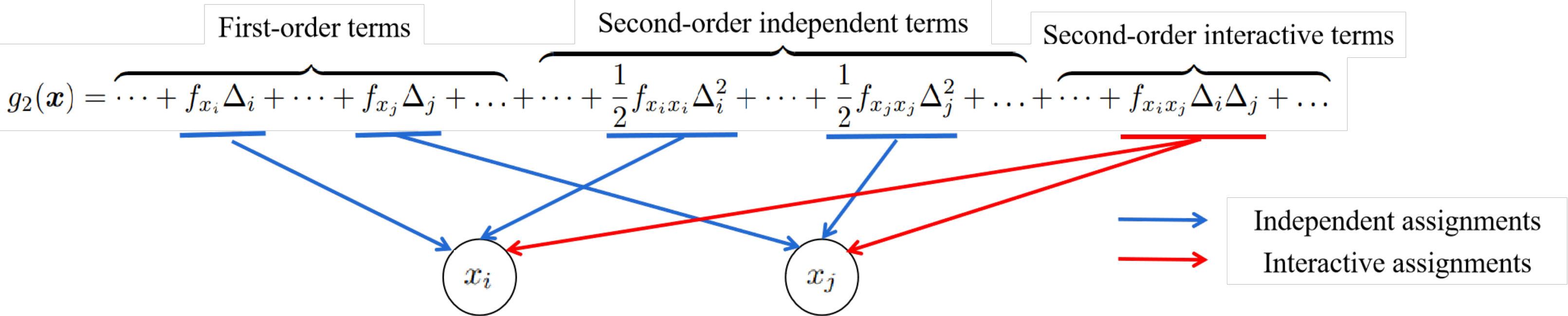}
\caption{An overview of Taylor attribution framework. Taking a second-order Taylor expansion as an example, $g_2(\bm{x})$ is composed of first-order, second-order independent, and second-order interactive terms. The first-order and second-order independent terms of $x_i$ can be clearly assigned to $x_i$, as shown in blue line. The second-order interactive term between $x_i$ and $x_j$ should be and only be assigned to $x_i$ and $x_j$, as shown in red line. }
\label{Taylor attribution diagram}
\end{figure*}

The Taylor expansion of $f$ expanded at sample $\bm{x}$ is\footnote{Noted that although the deep relu network is not differentiable such that Taylor expansion is not applicable, we can use networks with softplus activation (approximation of relu) to provide an insight to the rationale behind relu net.},
\begin{center}
$ f(\bm{\tilde x}) - f(\bm{x})  =  g_K(\bm{x},\Delta) + \epsilon(\bm{\tilde x}, K),$
\end{center}
where $g_K(\bm{x},\Delta)$ is the $K$-order Taylor expansion function of $f$, $\bm{\tilde x}$ denotes a baseline point which acts as a ``refernce'' state, vector $\Delta \coloneqq \bm{\tilde x} - \bm{x}$, and $\epsilon(\bm{\tilde x}, K)$ is the approximation error between $f(\bm{x})$ and $g_K(\bm{x},\Delta)$ at point $\bm{\tilde x}$.
The left side of equation, $f(\bm{\tilde x}) - f(\bm{x})$, represents the output change, which can be considered as the effect of input change $\Delta$.
The attribution problem becomes to decompose the effect to each $\Delta_i \coloneqq \tilde x_i - x_i$, the change of feature $i$.
It's difficult to decompose directly the effect due to the complexity of $f$.
As $g_K(\bm{x},\Delta)$ is an approximation of the output change, we instead decompose $g_K(\bm{x},\Delta)$ into a attribution vector $\bm{a} = [a_1, \dots, a_n]^{\rm T}$, where $a_i$ denotes the attribution score of feature $x_i$. An overview of Taylor attribution framework is illustrated in Figure \ref{Taylor attribution diagram}. For convenience, we summary the descriptions of main symbols in this paper in Table \ref{tab:my_label}.

\subsection{First-order Taylor Attribution}
The first-order Taylor expansion function $g_1(\bm{x}, \Delta)$ is
\begin{center}
$g_1(\bm{x}, \Delta) =  f_{\bm{x}}^\mathrm{T} \Delta = \sum_{i} f_{x_i} \Delta_i,$
\end{center}
where $f_{x_i}$ denotes the derivative of $f$ with respect to $x_i$. The linear approximation function in first-order Taylor expansion, $g_1(\bm{x}, \Delta)$, is additive across features and can be easily decomposed. It is obvious that $f_{x_i} \Delta_i$ quantifies the contribution of feature $x_i$, i.e.,
\begin{center}
$a_i = f_{x_i} \Delta_i$.
\end{center}

\subsection{Second-order Taylor Attribution}
The second-order Taylor expansion has a smaller approximation error $\epsilon$ than the first-order one, so that it is expected more faithful to the model $f$.
The second-order Taylor expansion function $g_2(\bm{x}, \Delta)$ is given by
\begin{center}
$g_2(\bm{x}, \Delta)=  \underbrace{f_{\bm{x}}^\mathrm{T} \Delta}_{\text{$T^\alpha$}} + \underbrace{\frac{1}{2}\Delta^T H_{\bm{x}} \Delta}_{\text{$T^\beta$}},$
\end{center}
where $H_{\bm{x}}$ is the Hessian matrix, i.e., second-order partial derivative matrix, of $f$ at $\bm{x}$. We denote the first-order and second-order Taylor terms as $T^\alpha$ and $T^\beta$, respectively.

The second-order Taylor expansion function $g_2(\bm{x}, \Delta)$ is indistinct in determining feature contributions compared with first-order one due to the Hessian matrix. To make the attribution more clear, we decompose $H_{\bm{x}}$ into two matrices, an independent matrix $H_{\bm{x}}^{d}$ and an interactive matrix $H_{\bm{x}}^{t} \coloneqq H_{\bm{x}} - H_{\bm{x}}^{d}$. Here $H_{\bm{x}}^{d}$ is a diagonal matrix composed of the diagonal elements in $H_{\bm{x}}$, which describes the second-order isolated effect of features, and $H_{\bm{x}}^{t}$ represents the interactive effect between features.
$g_2(\bm{x}, \Delta)$ could be rewritten as the sum of first order terms $T^\alpha$, second-order independent terms $T^{\beta_d}$, and second-order interactive terms $T^{\beta_t}$,
\begin{equation}\label{Second-order Taylor decomposition decomposition} \nonumber
 g_2(\bm{x}, \Delta) = \underbrace{f_{\bm{x}} \Delta}_{\text{$T^\alpha$}} + \underbrace{\frac{1}{2}  \Delta ^T H_{\bm{x}}^{d} \Delta}_{\text{$T^{\beta_d}$}}
 + \underbrace{\frac{1}{2}\Delta^T H_{\bm{x}}^{t} \Delta}_{\text{$T^{\beta_t}$}}.
\end{equation}

Accordingly, the attribution of  $g_2(\bm{x}, \Delta)$ to $x_i$ should be
\begin{center}
    $a_i = a_i^\alpha +a_i^{\beta_d} + a_i^{\beta_t}$,
\end{center}
where $a_i^{\alpha}, a_i^{\beta_d}$ and $a_i^{\beta_t}$ represent the assigned contributions from $T^{\alpha}$, $T^{\beta_d}$ and $T^{\beta_t}$, respectively.
The contributions from independent terms $T^{\alpha}$ and $T^{\beta_d}$ can be clearly identified as
\begin{center}
    $a_i^{\alpha} = T^{\alpha}_i = f_{x_i}\Delta_i, \quad a_i^{\beta_d} = T_i^{\beta_d} = \frac{1}{2} f_{x_ix_i}\Delta_i^2$,
\end{center}
where $T^{\alpha}_i$ and $T^{\beta_d}_i$ denote the first-order terms and second-order independent terms of feature $x_i$, respectively.

The difficulty lies on how to assign the contribution from interactive terms $T^{\beta_t}$. We propose to handle it by following an intuition behind: the attribution $a_i^{\beta_t}$ is from $T^{\beta_t}$ and should be the sum of assignments from each interactive effect involving feature $x_i$,
\begin{center}
    $a_i^{\beta_t} = \sum_{j\neq i} a_{i,\{x_i,x_j\}}^{\beta_t}
    =  \sum_{j \neq i} w_{i}^{\{i,j\}} T^{\beta_t}_{\{x_i,x_j\}}$,
\end{center}
where $T^{\beta_t}_{\{x_i,x_j\}} = f_{x_ix_j}\Delta_i\Delta_j$ denotes the second-order interactive terms corresponding to feature $x_i$ and $x_j$, weight $w_{i}^{\{i,j\}}$ characterizes the assignment of the interactive terms to $x_i$, and $a_{i,\{x_i,x_j\}}^{\beta_t}$ is the attribution from  $T^{\beta_t}_{\{x_i,x_j\}}$.

The determination of the assignment weight $w_{i}^{\{i,j\}}$ is complicated and depends on specific case.
However, it's considered that \textit{the interactive terms of two features should be only attributed to these two features}.
Consider the interactive terms between $x_i$ and $x_j$, the assignment should satisfy $a_{i,\{x_i,x_j\}}^{\beta_t} + a_{j,\{x_i,x_j\}}^{\beta_t} = T^{\beta_t}_{\{x_i,x_j\}}$, i.e., $w_{i}^{\{i,j\}} + w_{j}^{\{i,j\}} = 1$.
The second-order interactive term are equally assigned to  $x_i$ and $x_j$, i.e., $w_{i}^{\{i,j\}} =w_{j}^{\{i,j\}} =\frac{1}{2}$ in Integrated Gradients \cite{sundararajan2017axiomatic}, as shown in the reformulation in Section 3.1.



\subsection{High-order Taylor Attribution}
The analysis on second-order expansion can be naturally extended to high-order expansion where $K > 2$.
Let $T^{\gamma}$ denote all high-order expansion terms, including second-order expansion terms.
The high-order Taylor expansion function is
\begin{center}
$g_K(\bm{x},\Delta) = T^{\alpha} + T^{\gamma_d}  + T^{\gamma_t},$
\end{center}
where $T^{\gamma_d}$ and $T^{\gamma_t}$ denote high-order independent and interactive terms, respectively.

Analogously to the second-order case, the attribution of feature $x_i$ in high-order expansion is given by
\begin{center}
    $a_i = a_i^{\alpha} + a_i^{\gamma_d} + a_i^{\gamma_t}$,
\end{center}
where $a_i^{\gamma_d}$, $a_i^{\gamma_t}$ represent the assigned contributions from $T^{\gamma_d}$ and $ T^{\gamma_t}$, respectively.  The attribution from first-order term and high-order independent term is clear,
\begin{center}
    $a_i^{\alpha} = T_i^{\alpha}$, \quad $a_i^{\gamma_d} = T_i^{\gamma_d}$,
\end{center}
where $T^{\gamma_d}_i$ represent the high-order independent terms of feature $x_i$.
The attribution from interactive terms, $a_i^{\gamma_t}$, consists of all assignments from interactive terms involving $x_i$,
\begin{center}
$a_i^{\gamma_t} = \sum_{A} a_{i,A}^{\gamma_t}$,  $x_i \in A$,
\end{center}
where $a_{i,A}^{\gamma_t}$ denotes the attribution from interactive terms corresponding to features in the feature subset $A$.
Note that interactive terms $T^{\gamma_t}_A$ should be only assigned to the features in the subset $A$, i.e., $\sum_{i \in A} a_{i,A}^{\gamma_t} = T^{\gamma_t}_A$.

\subsection{The Selection of Baseline Point}
From the Taylor attribution framework, the attribution of feature $x_i$ could be seen as a polynomial function of $\Delta_i$ (i.e., $\tilde x_i - x_i$), and hence it highly depends on $\tilde x_i$.  Given a baseline of constant vector $\bm{\tilde x} = \bm{c}$ as many attribution methods did,  the attribution of feature whose value is far from $c$ may be overestimated due to a large $\Delta_i$, while the attribution of feature whose value is close to $c$ may be underestimated even if it is important to the decision making process.
Such different attributions are a bias in many tasks. For example, in image classification, it's unreasonable to attribute according to the value of features (i.e., pixel values). Specifically, given a black image as baseline, pixels in white color have a large $\Delta_i$ close to $255$, while pixels in black color have a small $\Delta_i$ close to $0$. Correspondingly, the attribution methods will biasedly highlight white pixels while neglecting black pixels even if black pixels make up the object of interest. Hence the selection of baseline point $\bm{\tilde x}$ plays a significant role.

Baseline point is used to represent an ``absence" of a feature, by which the attribution methods calculate how much the output of the model would decrease considering the absence of the feature \cite{sturmfels2020visualizing}.
To avoid incorporating aforementioned bias into the attribution process, attribution methods should choose a unbiased baseline which satisfies there is no big differences among $\Delta_i$ of different features. That is, $\Delta_i$ should be similar to $\Delta_j$ for random two feature dimensions. One option is setting $\Delta$ as a constant vector $\bm{c}$, and its corresponding baseline is $\bm{\tilde x} = \bm{x} + \bm{c}$. Such baselines indeed solve the bias issue, however they usually don't make a difference to the output of the model.  Another alternative is to neutralize the bias by averaging multiple baselines whose $\Delta$s are sampled from some distributions with small variance, such as uniform and Gaussian distributions \cite{smilkov2017smoothgrad}. Noted that it's difficult for neutralization over samples from a distribution with large variance. This may explain why SmoothGrad and Integrated Gradients will success with small Gaussian noise level while fail with large noise level.

\section{Revisiting Existing Attribution Methods}
The proposed Taylor attribution framework is very general to unify most existing attribution
methods. On one hand, most mainstream attribution methods aim to assign or decompose the output difference between the input sample and the baseline, $\Delta f$ = $f(\bm{x})-f(\tilde{\bm{x}})$, to each input feature, which can be thought of a function of $\Delta f$. On the other hand, the Taylor expansion adopted in the framework could decompose $\Delta f$  into the sum of input features’ effects (Taylor terms). Therefore, the attribution can be unified into our framework, i.e., the attribution could be reformulated as a function of the Taylor terms.

In the following, we first reformulate seven mainstream attribution methods into the proposed framework, and then systematically analyze their rationale, fidelity, and limitations based on the reformulations.
Finally, we establish three principles for a good attribution.

\subsection{Unified Reformulations}
We reformulate seven mainstream attribution methods into the proposed Taylor attribution framework, which are
Gradient*Input \cite{shrikumar2016not},
Occlusion-1 \cite{zeiler2014visualizing},
Occlusion-patch \cite{zintgraf2017visualizing},
DeepLIFT (Rescale) \cite{shrikumar2017learning},
$\epsilon$-LRP \cite{bach2015pixel},
Integrated Gradients \cite{sundararajan2017axiomatic} and Expected Gradients \cite{erion2019learning}.
In the following, we will briefly introduce these attribution methods, and then present their reformulations.
We also discuss several popular attribution methods that cannot be unified into the framework.
Note that the expansion point is the input sample $\bm{x}$ if not specified.  All the proofs of theorems are in the Supplementary materials \footnote{The supplementary materials could be downloaded at:\\
https://arxiv.org/abs/2008.09695}.

\subsubsection{Gradient*Input.} The attribution in Gradient*Input is calculated by multiplying the partial derivatives (of the output to the input) with the input, i.e., $a_i = f_{x_i}(\bm{x}) x_i$.

\noindent
\textbf{Theorem 1.}
\emph{Gradient*Input can be reformulated as a first-order Taylor attribution w.r.t the baseline point $\bm{\tilde x} = \bm{0}$,}
\begin{equation}\nonumber
a_i = T^{\alpha}_i.
\end{equation}

That is, the attribution of $x_i$ in Gradient*Input is $f_{x_i}\Delta_i$.

\subsubsection{Occlusion-1.} Occlusion-1 attribution calculates how much the prediction changes induced by occluding feature $x_i$ with a zero baseline. The new occluded input is written as $\bm{x}|_{x_i = 0}$.  Then the attribution of feature $x_i$ is defined as the change of the output, $a_i = f(\bm{x}) - f(\bm{x}|_{x_i = 0})$.

\noindent
\textbf{Theorem 2.}
\emph{
The attribution of $x_i$ in Occlusion-1 can be reformulated as the sum of first-order and high-order independent terms of $x_i$ at baseline point $\bm{\tilde x} =  \bm{x}|_{x_i = 0}$,}
\begin{equation} \nonumber
a_i = T^{\alpha}_i + T^{\gamma_d}_i.
\end{equation}

The attribution of $x_i$ in Occlusion-1 is $f_{x_i}\Delta_i +  \frac{1}{2}f_{x_ix_i}\Delta_i^2$ in the second-order Taylor attribution.

\subsubsection{Occlusion-patch.}
The attribution in Occlusion-patch (Occlusion-p for short) is conducted on a patch level. It constructs a zero patch baseline $\bm{x}|_{\bm{p_i} = \bm{0}}$ by occluding an image patch $\bm{p_j}$, and defines the prediction change $f(\bm{x}) - f(\bm{x}|_{\bm{p_j} = \bm{0}})$ as the attribution of feature in $\bm{p_j}$.

\noindent
\textbf{Theorem 3.}
\emph{The attribution of $x_i \in \bm{p_j}$ in Occlusion-p can be reformulated as the sum of first-order, high-order independent terms of features in patch $\bm{p_j}$, and all high-order interactive terms involving the features in patch $\bm{p_j}$,}
\begin{equation} \nonumber
\begin{aligned}
a_i & = T_{\bm{p_j}}^{\alpha} + T^{\gamma_d}_{\bm{p_j}} + T^{\gamma_t}_{\bm{p_j}} \\
& = \sum\nolimits_{x_i \in \bm{p_j}}  T_i^{\alpha} + \sum\nolimits_{x_i \in \bm{p_j}} T^{\gamma_d}_i +  \sum\nolimits_{A \subset \bm{p_j}} T^{\gamma_t}_A.
\end{aligned}
\end{equation}

Particularly, the $a_i$ is $\sum_{i \in \bm{p_j}} f_{x_i}\Delta_i + \sum_{i \in \bm{p_j}}\frac{1}{2}f_{x_ix_i}\Delta_i^2 + \sum_{i \in \bm{p_j}}\sum_{j \in \bm{p_j}} f_{x_ix_j}\Delta_i \Delta_j$ in second-order setting.

\subsubsection{DeepLIFT and $\epsilon$-LRP.}
DeepLIFT and $\epsilon$-LRP compute relevance scores by using a recursive relevance propagation in a layer-wise manner. In DeepLIFT Rescale rule, $x_i^{(l)}$ and $x_j^{(l+1)}$ denote the neuron $i$ at $l$-th layer and the neuron $j$ at $(l+1)$-th layer, respectively, and $x_j^{(l+1)} = \sigma(\sum_i w_{ji}x_i^{(l)} + b_{j})$.
Here $w_{ji}$ is the weight parameter, $b_{j}$ is the additive bias, and $\sigma$ is a non-linear activation function.
DeepLIFT propagates the output difference between an input $\bm{x}$ and a baseline $\bm{\tilde x}$ to the input layer, and it calculates the relevance score of $x^{(l)}_i$ to $x^{(l+1)}_j$, denoted as $a^{(l)}_{ij}$, by
\begin{equation}\nonumber
a^{(l)}_{ij}  =  \frac{z_{ji}^{(l)} - \tilde z_{ji}^{(l)}}{\sum_{i'} z_{ji'}^{(l)} - \sum_{i'} \tilde z_{ji'}^{(l)}} a^{(l+1)}_j,
\end{equation}
where $z^{(l)}_{ji}$ = $w_{ji}x_i^{(l)}$ is the weighted impact of  $x_i^{(l)}$ to $x_j^{(l+1)}$,
analogously $\tilde z_{ji}^{(l)}$ = $w_{ji}\tilde x_i^{(l)}$ denotes the weighted impact of the baseline, and $a^{(l+1)}_j$ = $\sum_k a^{(l+1)}_{jk}$ denotes the total relevance score of neuron $x^{(l+1)}_j$ to all neurons in $(l+1)$-th layer.
The formula of $\epsilon$-LRP is similar to DeepLIFT, please see details in the supplementary materials.

\noindent
\textbf{Theorem 4.}
\emph{The relevance score of $x^{(l)}_i$ to $x^{(l+1)}_j$ in DeepLIFT can be reformulated as the weighted sum of first-order term of $x^{(l)}_i$ and all high-order terms at baseline $\tilde x_i^{(l)}$,}
\begin{equation} \nonumber
a_{ij}^{(l)} = T_i^{\alpha} + \frac{z_{ji}^{(l)} - \overline z_{ji}^{(l)}}{\sum_{i'} z_{ji}^{(l)} - \sum_{i'} \overline z_{ji}^{(l)}} T^{\gamma}.
\end{equation}

In the second-order setting, the attribution $a_{ij}^{(l)}$ is $f_{x_i}\Delta_i +  \frac{z_{ji}^{(l)} - \overline z_{ji}^{(l)}}{\sum_{i'} z_{ji}^{(l)} - \sum_{i'} \overline z_{ji}^{(l)}} (\sum_{i} \frac{1}{2}f_{x_ix_i}\Delta_i^2 + \sum_{ij}  f_{x_ix_j}\Delta_i \Delta_j$).

\subsubsection{Integrated Gradients.}
The attribution in Integrated Gradients integrates the gradients along the straight line path from a baseline point $\bm{\tilde x}$ to an input $\bm{x}$.
The points along the path are denoted as $\bm{x'}$ = $\bm{\tilde x} + \alpha( \bm{x} - \bm{\tilde x}), \alpha \in [0,1]$. The attribution of feature $x_i$ is computed by
\begin{equation}\label{Integrated Gradients}
a_{i} = ( x_i - \tilde x_i) \int_0^1 \frac{\partial {f(\bm{\tilde x} + \alpha( \bm{x} - \bm{\tilde x}))}}{\partial {x_i}} d\alpha.
\end{equation}

\noindent
\textbf{Theorem 5.}
\emph{ The attribution of $x_i$ in Integrated Gradients can be reformulated as the sum of first-order term of $x_i$, high-order independent terms of $x_i$, and an assignment from high-order interactive terms involving $x_i$ at baseline $\bm{\tilde x}$,}
\begin{equation} \nonumber
a_i = T_i^{\alpha} + T_i^{\gamma_d} + a^{\gamma_t}_i,
\end{equation}
\emph{where $a^{\gamma_t}_i$ = $\sum_{K = 2}^{\infty} \sum_{k_i} \frac{k_i}{K}  ( \Delta_i^{k_i}  \prod_{\sum_j k_j = K-k_i} C \Delta_j^{k_j})$ is the assignment, and $C$ = $\frac{1}{K!}\tbinom{K}{k_1,\dots,k_n}\frac{\partial f(\bm{x})} {\partial x_1^{k_1}\dots \partial x_i^{k_i} \dots \partial x_n^{k_n} }$ is the Taylor expansion coefficient of $\Delta_1 ^{k_1}\dots \Delta_i ^{k_i}\dots \Delta_n^{k_n}$.}

In brief, Integrated Gradients allocates $\frac{k_i}{K}$ proportion of the high-order interactive term $\Delta_1 ^{k_1}\dots \Delta_i ^{k_i}\dots \Delta_n^{k_n} $ to $x_i$.
In the second-order setting, the attribution of $x_i$ in Integrated Gradients is $f_{x_i}\Delta_i + \frac{1}{2}f_{x_ix_i}\Delta_i^2 + \frac{1}{2}\sum_{j\neq i} f_{x_ix_j}\Delta_i\Delta_j$.

\subsubsection{Expected Gradients.}
Expected Gradients is an extension of Integrated Gradients.
Expected Gradients samples baseline points from a prior distribution $p_D(\bm{\tilde x})$, instead of specifying only one baseline point in Integrated Gradients.
The attribution $a_i$ is then computed by integrating the Integrated gradients attributions along the baseline distribution,
\begin{equation}\label{Expected Gradients}
a_{i} =  \int_{\bm{\tilde x}}  p_D(\bm{\tilde x}) ( x_i - \tilde x_i)   \int_0^1 \frac{\partial {f(\bm{\tilde x} + \alpha( \bm{x} - \bm{\tilde x}))}}{\partial {x_i}} d\alpha d\bm{\tilde x}.
\end{equation}
A common choice for prior distribution is to add a zero-mean, independent Gaussian distribution to $\bm{x}$. That is, $\bm{\tilde x} \sim N(\bm{x},\sigma^2)$.

We can see the attribution in Expected Gradients, Eq. \ref{Expected Gradients}, is an integral over the baselines distribution of the attribution in Integrated Gradients, Eq. \ref{Integrated Gradients}.
Therefore, we can have a direct corollary that the attribution in Expected Gradients can also be  reformulated into the Taylor attribution framework, and its reformulation is a integral of the reformulation of Integrated Gradients in theorem 5. Please see the specific reformulation and derivation in the supplementary materials.

In additional to the seven attribution methods, some attribution methods have been proven that they are the first-order Taylor attributions at (well-chosen) nearest root point, such as LRP-$\alpha\beta$ \cite{montavon2019layer} and DeepTaylor \cite{montavon2017explaining}.
In the mean time, there are also several popular attribution methods that cannot be unified into the Taylor attribution framework, e.g.,
Deconvnet \cite{zeiler2014visualizing} and Guided BP \cite{springenberg2014striving}.
The two attribution methods do not analyze the output change $\Delta f$ = $f(\bm{x})-f(\tilde{\bm{x}})$ as the proposed framework does.
It has been theoretically and  empirically shown they are to recover the input \cite{adebayo2018sanity, nie2018theoretical}.

\subsection{Theoretical Analysis of the Attribution Methods}
The proposed Taylor reformulations enable us to examine rationales, measure fidelity, and analyze limitation for the attribution methods in a systematic and theoretical way.
Firstly, we find \textbf{Gradient*Input}, \textbf{Occlusion-1}, and \textbf{Occlusion-patch} have a large approximation error $\epsilon$ (i.e., low fidelity) as they all fail to completely reflect the high-order Taylor terms.
Theorem 1 shows Gradient*Input is a first-order Taylor attribution, which only takes the first-order terms into consideration.
Although Occlusion-1 partially characterizes the high-order independent effects in Theorem 2, it fails to attribute the interactive terms. The complex interactions among features (pixels) always contains critical information for prediction in DNNs.
Theorem 3 shows Occlusion-patch considers the overall effects of the features in the patch, including both independent and interactive terms. However, it assigns the same contribution score to all features in the patch, which fails to provide fine-grained attributions. Moreover, the interactive effects among different patches are neglected in Occlusion-patch.

From the reformulations, we can see \textbf{DeepLIFT}, $\epsilon$\textbf{-LRP}, and \textbf{Integrated Gradients} have a small approximation error $\epsilon$ as they attribute the high-order terms.
DeepLIFT and $\epsilon$-LRP assigns weighted high-order terms of all features to feature $x_i$, which obviously fails to distinguish the high-order contributions of different features.
As DeepLIFT and $\epsilon$-LRP conduct in a layer-wise manner, the impact of high-order terms in each layer is much smaller than the one in a network, which may relieve the incorrect assignment problem.

Integrated Gradients is an average of first-order derivatives along the path.
However, Theorem 5 shows that Integrated Gradients not only attributes the first and  high-order independent terms, but also correctly assigns the interactive terms.
Hence it's considered that Integrated Gradients is a superior attribution.
This theoretical finding may provide an insight into why Integrated Gradients can well identify important features in input image.
The performance of Integrated Gradients highly depends on baseline. However, it is difficult to select a good baseline. If use a  black image as a baseline, integrated gradients will not highlight black pixels as important even if they make up the object of interest.

Lastly, we discuss about \textbf{Expected Gradients}. Expected Gradients relieved the problem induced by baseline by averaging among multiple baselines sampled from a Gaussian distribution, as analyzed in section 2.4. It reduces the probability that the attribution is dominated by a specific baseline.

\subsection{Three Principles of Attribution}
Based on the reformulations and analysis, we find a good Taylor attribution depends on three key factors: i) the Taylor approximation error $\epsilon(\bm{\tilde x}, K)$;
ii) whether the Taylor terms in $g_K(\bm{x},\Delta)$ are assigned correctly;
iii) the baseline point $\tilde{\bm{x}}$.
Accordingly, we establish three principles of a good attribution and advocate the principles should be followed by other attribution methods.

\noindent
\textbf{First principle:}
After Taylor reformulating, an attribution method should \emph{has a low approximation error} $\epsilon(\tilde{\bm{x}}, K)$, $\forall \bm{\tilde x}$. This principle is similar to the completeness axiom.

\noindent
\textbf{Second principle:}
After Taylor reformulating, an attribution method should  \emph{correctly assign the independent and interactive terms}.
For instance, the first-order and high-order independent terms of feature $x_i$ should only be assigned to $a_i$, and the interactive terms of features in subset $A$ should only be attributed to the features in subset $A$.

\noindent
\textbf{Third principle:}
After Taylor reformulating, an attribution method should choose \emph{an unbiased baseline}.

Table \ref{A summary of the principles} presents a summary of the principles followed by the attribution methods, in which Occlusion-1 and Occlusion-p partially follow the second principle because they partially attribute high-order terms, as discussed above.

\begin{table}[t]
\centering
\renewcommand\arraystretch{1.5}
\begin{tabular}{ m{2.8cm}|C{1.25cm}|C{1.25cm}|C{1.25cm}} \hline
\textbf{Principles} &  \textbf{First} & \textbf{Second}  & \textbf{Third} \\  \hline
Gradient*Input &  &   &  \\ \hline
Occlusion-1 \& -p &  & partially & \\ \hline
DeepLIFT \& $\epsilon$-LRP & $\surd$ & & \\ \hline
Integrated & $\surd$ & $\surd$ & \\ \hline
Expected  & $\surd$ & $\surd$ & $\surd$\\ \hline
\end{tabular} \\
\caption{A summary of the principles followed by the attribution methods. }
\label{A summary of the principles}
\end{table}

\section{Experiments}
In this section, we empirically validate our Taylor reformulations, and then investigate the relationship between attribution performance and the three principles via benchmarking on MNIST and Imagenet.
We use GI, Occ-1, Occ-p, DL, IG, and  EG to denote Gradient*Input, Occlusion-1, Occlusion-p, DeepLIFT Rescale, Integrated Gradients, and Expected Gradients, respectively.

\subsection{Empirical Validations of Reformulations}
In this section, we empirically validate our Taylor reformulations by comparing the attribution results from the original attribution methods and the corresponding Taylor reformulations.
We firstly compute the attribution vector by the original attribution method and their Taylor reformulation. Then we adopt average percentage change as the metric to measure the difference between the attributions,
\begin{center}
    $d = \frac{1}{N}\sum_{i = 1}^N\frac{||\bm{a}_o(i) - \bm{a}_r(i)||^2}{||\bm{a}_o(i)||^2} \times 100 \%,$
\end{center}
where $\bm{a}_o(i)$ and $\bm{a}_r(i)$ denote the attribution vectors  obtained by the original attribution method and the corresponding Taylor reformulation, respectively, for sample $i$. And $N$ is the number of samples in the dataset.
As DeepLIFT attributes in a layer-wise manner, we compute the percentage change of each layer and then average them.

We conduct the validation experiments on three models: $\quad$ i) \textbf{Poly}, a second-order polynomial model. ii) \textbf{M-sg}, a three-layer multi-layer perceptron (MLP) model with sigmoid activation, iii) \textbf{C-sg}, a three-layer CNN model with sigmoid activation. These models are all trained on MNIST \footnote{http://yann. lecun. com/exdb/mnist/} dataset. The average percentage change metrics are averaged on 3k validation set.
We use second-order Taylor reformulations to validate the theories as higher-order one is always computationally intractable. Our theoretical results would expect the average percentage change metric should be small for the models which could be well approximated by second-order Taylor expansion.

\begin{table}[t]
\centering
\renewcommand\arraystretch{1.5}
\begin{tabular}{ m{1.5cm}|C{1.25cm}|C{1.25cm}|C{1.25cm}} \hline
Models & Poly & M-sg  & C-sg\\  \hline
GI & 0 & 0  & 0\\ \hline
Occ-1& 0 &  0.05\%   & 13.17\% \\ \hline
Occ-2$\times$2 & 0 & 0.35\%   & 21.22\%\\ \hline
Occ-4$\times$4 & 0 & 2.00\%   & 35.93\%\\ \hline
DL & 0 & 2.35\%  & 32.76\%\\ \hline
IG & 0.12\% & 3.50\%   & 40.29\% \\ \hline
\end{tabular} \\
\caption{Average percentage changes between original attribution methods and their Taylor reformulations. }
\label{Differences between formula and reformula.}
\end{table}

Table \ref{Differences between formula and reformula.} lists the average percentage change metrics of the aforementioned attribution methods. Occlusion-p is implemented for patch size 2$\times$2 and 4$\times$4. It can be seen that the metrics between attribution methods and their Taylor reformulations are equal to $0$ in second-order Poly model (The marginal difference of Integrated Gradients is due to the integral approximation error) and M-sg models. This demonstrates the correctness of  our theoretical reformulations.

The metrics on C-sg model are obviously larger than other models.
We find that the discrepancy is mainly due to the incapability of second-order Taylor expansion to approximate C-sg model, instead of the proposed reformulations. We compute that the average (normalized) approximation error $\bar \epsilon = \frac{1}{N}\sum_i |\epsilon_i|/|\Delta f_i|$ of second-order Taylor expansion to Poly, M-sg, and C-sg model are 0, 0.084, 0.420 respectively.  We observe that C-sg model has the largest approximation error. Moreover, there is a strong correlation between the percentage change and approximation error, which implies that the percentage change is resulted by the approximation error. For example, the correlation coefficient for IG method on C-sg model is $0.72$.

\subsection{Attribution Assessment}
To investigate the relationship between attribution performance and three principles, we benchmark the six attribution methods in terms of infidelity and object localization accuracy on MNIST and Imagenet \cite{ILSVRC15} \footnote{$\epsilon$-LRP has been shown equivalent to GI theoretically, so we do not show the results of $\epsilon$-LRP.}.

\begin{figure}[t]\centering
\includegraphics[scale=0.33]{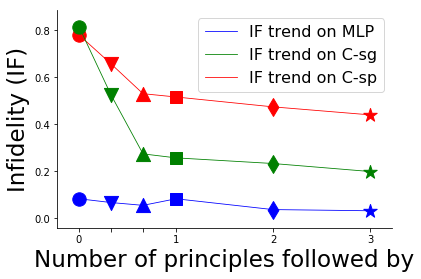}
\caption{Infidelity trends on MLP (blue), C-sg (green), and C-sp models (red).
The circle, triangle down, triangle up, square, diamond, and star denotes GI, Occ-1, Occ-p, DL, IG, and EG, respectively.
$x$-axis denotes the number of principles the methods followed by, and $y$-axis denotes infidelity. }
\label{IFtrend}
\end{figure}

We evaluate the infidelity of these attribution methods on images from MNIST dataset. We adopt the infidelity metric proposed in \cite{yeh2019fidelity}, which quantifies the degree to which it captures how the predictor function itself changes in response to significant perturbations. We use the square removal perturbation to assess the attributions of MLP, C-sg, and C-sp (CNN softplus) models. The infidelity trends (Figure \ref{IFtrend}) show a negative correlation between the infidelity and the number of principles followed by these methods.
Furthermore, we also compare the infidelity of first and second-order Taylor explanations on MNIST dataset. Experimental results show that the average infidelities of the second-order one (0.27 on C-sg, 0.75 on C-sp) are significantly lower than the infidelity of the first-order one (0.48 on C-sg, 0.84 on C-sp). The results indicate that additional high-order Taylor terms indeed help improve the explanation.

\begin{figure}[t]\centering
\includegraphics[scale=0.33]{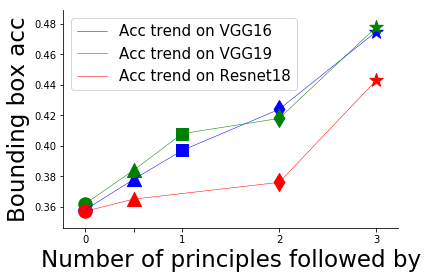}
\caption{Bounding box accuracy trends on VGG16 (blue), VGG19 (green), and Resnet18 models (red). 
$x$-axis denotes the number of principles followed by, and $y$-axis denotes Bounding box accuracy.}
\label{Acctrend}
\end{figure}

We also investigate the relationship by measuring the attribution performance of these attribution methods on Imagenet. Here bounding box accuracy \cite{Karl2020Restricting} is adopted to evaluate how well attribution methods locate objects of interest. Assume the annotated bounding box contains $n$ pixels. We select top $n$ pixels according to ranked attribution scores and count the number of pixels $m$ inside the bounding box. The ratio $\frac{m}{n}$ is used as the metric of localization accuracy.
We only consider the images whose bounding boxes cover less than 33\% of the input image. The bounding box accuracies are calculated on VGG16, VGG19, and Resnet18 networks.
The trends in Figure \ref{Acctrend} shows a positive correlation between the bounding box accuracy and the number of principles followed by these method.

\begin{figure}[t]\centering
\includegraphics[scale=0.3]{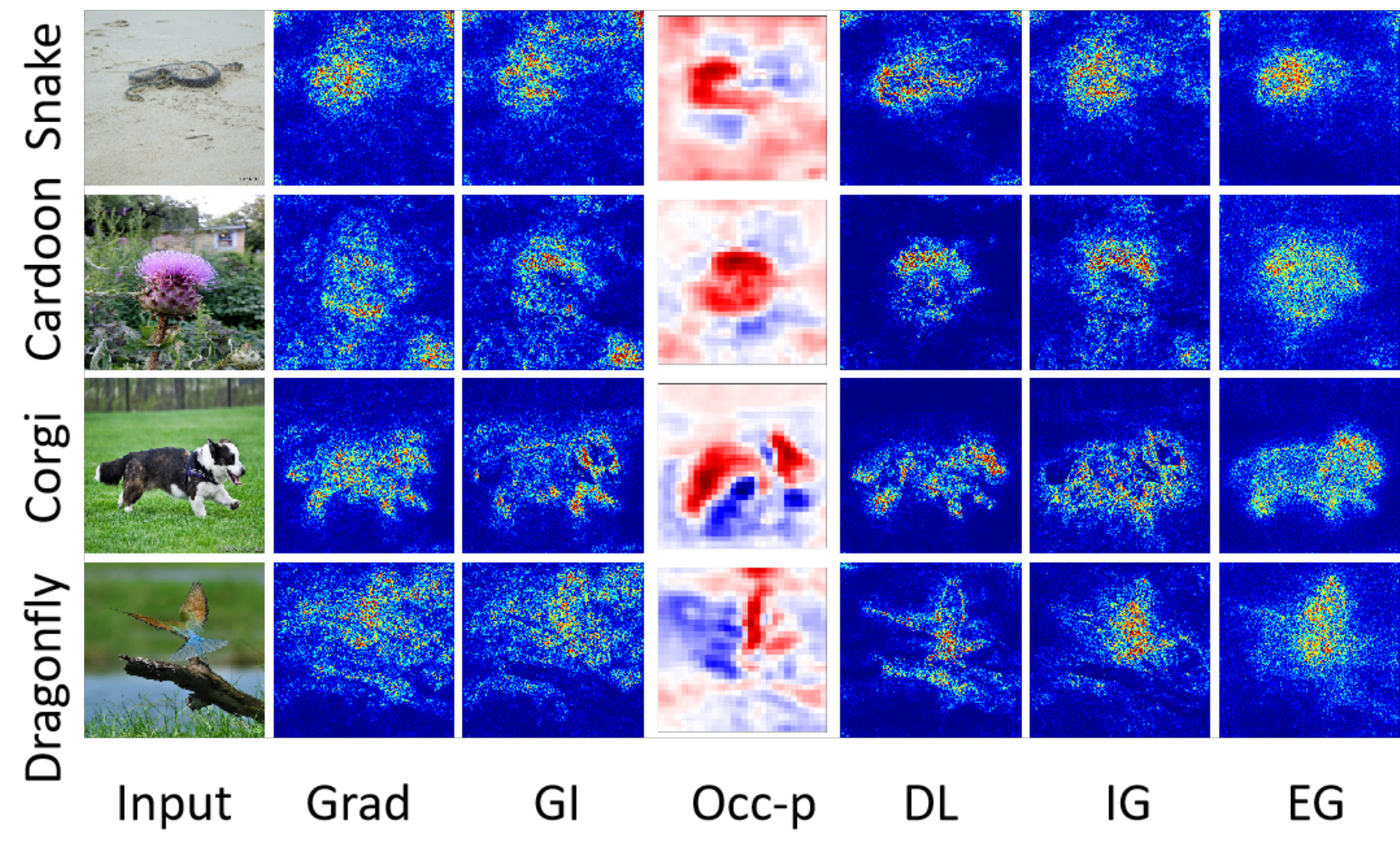}
\caption{Visualization saliency map comparison.}
\label{Visualization Comparison}
\end{figure}

The visualization comparisons among attribution methods are shown in Figure \ref{Visualization Comparison}. Firstly, the saliency maps based on Gradient and GI are visually noisy and involves some irrelevant regions to the prediction. Occ-p, which incorporates high-order terms into the attribution, can approximately highlight the object of interest. DL and IG accurately identify right location of object, however they are very sensitive to the selection of baseline. For example, When interpreting corgi image with a black image baseline, IG assigns significantly higher contributions to white areas than black areas. This is due to the bias induced by large differences of input change $\Delta$ among different feature dimensions. EG solved this issue, and its generated saliency maps, evenly distributed with less noises, are sharper, and clearly display the shapes and boarders of the objects.

\section{Related Work}
There are a few works on understanding the theoretical groundings of attribution methods.  Deconvnet and Guided BP have been theoretically proved \cite{nie2018theoretical} that they are essentially doing (partial) image recovery, which is unrelated to decision making. Some efforts have been devoted to unifying existing attribution methods recently. Several attribution methods are unified under the framework of additive feature attribution \cite{lundberg2017unified}, or reformulated as multiplying a modified gradient with input \cite{Marco2018Towards}
or summarized as first-order Taylor decomposition on different baseline points \cite{samek2020toward}. To our knowledge, this is the first work to unify these attribution methods and further analyze the interactive Taylor terms by high-order Taylor decomposition.

\section{Conclusion and Future Work}
In this work, we propose a general Taylor attribution framework and theoretically reformulate several mainstream attribution methods into the attribution framework. Based on reformulations, we systematically analyze the attribution methods in terms of their rationale, fidelity, and limitation. Based on the reformulation and analysis, we establish three principles for a good attribution. In the future work, we will promote a more general Taylor framework, i.e., unify more attribution methods that analyze or decompose the output difference into the proposed framework, such as Shapley value, DeepLIFT RevealCancel, and so on.

\section{Acknowledgements}
 The authors thank the anonymous reviewers for their helpful comments. This work is partially supported by the NSFC under grants Nos. 61673018, 61272338, 61703443 and Guangzhou Science and Technology Founding Committee under grant No.201804010255 and Guangdong Province Key Laboratory of Computer Science.

\bibliography{aaai21}

\clearpage
\appendix
\section{Appendix A: Proof of Theorems}
\noindent
\textbf{Proof of Theorem 1}
\begin{proof}
 The attribution of Gradient*Input $a_i = f_{x_i} x_i = f_{x_i}(x_i -0)$. It's obvious that $a_i$ corresponds to the first-order Taylor attribution w.r.t $\bm{\tilde x} = \bm{0}$.
\end{proof}

\noindent
\textbf{Proof of Theorem 2}
\begin{proof}
The attribution of feature $x_i$ is calculated by $a_i = f(\bm{x}) - f(\bm{x}|_{x_i=0})$. Here $\bm{x}|_{x_i = 0}$ denotes the baseline point that occludes feature $x_i$ of $\bm{x}$ while remains other features unchanged, and $f(\bm{x}|_{x_i=0})$ denotes the model output of the baseline. We firstly conduct Taylor expansion at point $\bm{x}$ w.r.t baseline point $\bm{x}|_{x_i = 0}$. It's observed that $\Delta_i = x_i $, while for $\forall j \neq i$, $\Delta_j = x_j - x_j = 0$.  By substituting $\Delta$ into Taylor expansion, we have, for $\forall j \neq i$, the first-order term $T^{\alpha}_j$ and all high-order independent terms of feature $x_j$  $T^{\gamma_d}_j$ equal to $0$. Moreover, all high-order interactive terms $T^{\gamma_t}$ also equals to $0$. That is, \\
\begin{equation}\nonumber
\begin{aligned}
a_i  &=  f(\bm{x}) - f(\bm{x}|_{x_i=0}) \\
& = T_i^{\alpha} +T_i^{\gamma_d} + \sum_{j \neq i} (T_j^{\alpha} +T_j^{\gamma_d} )+ T^{\gamma_t} \\
& = T_i^{\alpha} +T_i^{\gamma_d}.
\end{aligned}
\end{equation}
The attribution of $x_i$ in Occlusion-1 can be reformulated as the sum of first-order and high-order independent terms of $x_i$ w.r.t baseline point $x_i = 0$.
\end{proof}

\noindent
\textbf{Proof of Theorem 3}
\begin{proof}
The attribution of feature $x_i$ in patch $\bm{p_j}$ is calculated by $a_i = f(\bm{x}) - f(\bm{x}|_{\bm{p_j}=0})$. Here $\bm{x}|_{\bm{p_j}=0}$ denotes the baseline point that occludes all features in patch $\bm{p_j}$ of $\bm{x}$ while remains other patches unchanged, and $f(\bm{x}|_{\bm{p_j}=0})$ denotes the model output of the baseline. Similar to the derivation of Occlusion-1, we conduct Taylor expansion at point $\bm{x}$ w.r.t baseline point $\bm{x}|_{\bm{p_j} = 0}$.
We can see that for all other patches $\forall k \neq j$,  $\Delta_{\bm{p_k}} = \bm{0}$. By substituting $\Delta$ into Taylor expansion, we have, for $\forall k \neq j$, all first-order terms of features in the patch $\bm{p_k}$ equal to $0$. That is, $\sum_{x_m \in \bm{p_k}}T^{\alpha}_m = 0$. Similarly, all high-order independent terms of features in the patch $\bm{p_k}$ equal to $0$. i.e., $\sum_{x_m \in \bm{p_k}}T^{\gamma_d}_m = 0$.  Moreover, interactive terms among these patches equal to $0$. In addition, interactive terms between patch $\bm{p_j}$ and other patches also equal to $0$. That is, $\sum_{B \subset X \setminus \bm{p_j} } T^{\gamma_t}_{B} = 0$ and $\sum_{C \subset X} T^{\gamma_t}_{C} = 0$. Here $C$ denotes a feature subset of $X$ which not only involves features in patch $\bm{p_j}$, but also involves features in other patches, e.g., patch $\bm{p_k}$.
Therefore,
\begin{equation}\nonumber
\begin{aligned}
a_i  &= f(\bm{x}) - f(\bm{x}|_{\bm{p_j}=0}) \\
& = \sum_{x_i \in \bm{p_j}} (T^{\alpha}_i + T^{\gamma_d}_i) + \sum_k \sum_{x_m \in \bm{p_k}} (T^{\alpha}_m + T^{\gamma_d}_m)  \\
& + \sum_{A \subset \bm{p_j}} T^{\gamma_t}_{A}  + \sum_{B \subset X \setminus \bm{p_j}} T^{\gamma_t}_{B} + \sum_{C} T^{\gamma_t}_{C} \\
& =  \sum_{x_i \in \bm{p_j}} (T^{\alpha}_i + T^{\gamma_d}_i) + \sum_{A \subset \bm{p_j}} T^{\gamma_t}_{A},
\end{aligned}
\end{equation}
where $A$ goes through all subsets of feature set $\bm{p_j}$. $T^{\gamma_t}_{A}$ involves the overall high-order interactive terms among the features in subset $A$.
\end{proof}

\begin{table}[]
\centering
\begin{tabular}{L{1.675cm}|L{6.35cm}}
 \textbf{Symbol} & \textbf{Description}   \\ \hline
$X$ & Feature set $\{x_1, \dots, x_n\}$ \\ \hline
$A,B,C$ & Feature subset of $X$ \\ \hline
$\bm{x}$ & Input vector $[x_1, \dots, x_n]^{\rm T}$  \\ \hline
$\bm{\tilde x}$ & Baseline point vector  \\ \hline
$\Delta$ & Input change, defined as $\bm{\tilde x} - \bm{x}$  \\ \hline
$f(\bm{x})$ & DNN model with input $\bm{x}$ \\ \hline
$T^\alpha, T^\beta, T^\gamma$ & Taylor first, second, and high-order terms \\ \hline
$T^{{\beta}_{d}},T^{\gamma_d}$ & Taylor second, high-order independent terms \\ \hline
$T^{{\beta}_{t}}, T^{{\gamma}_{t}}$ & Taylor second, high-order interactive terms \\ \hline
$a_i$ & Attribution of feature $x_i$ \\ \hline
$a_i^{\alpha}, a_i^{\beta_d}, a_i^{\beta_t}$ &  Attribution of feature $x_i$ from $T^\alpha,T^{\beta_d},T^{\beta_t}$  \\ \hline
$a_i^{{\gamma}_d},a_i^{{\gamma}_t}$ & Attribution of feature $x_i$  from $T^{\gamma_d},T^{\gamma_t}$  \\ \hline
\end{tabular}
\vspace{-2mm}
\caption{Symbol descriptions in supplementary materials.}
\label{tab:my_label}
\vspace{-3mm}
\end{table}

\noindent
\textbf{Proof of Theorem 4}
\begin{proof}
Noted that $x_j^{(l+1)} = \sigma(y_j^{(l)})$ (here $y_j^{(l)} = \sum_{i'} z_{ji}^{(l)}+b_j$) is the $j$-th neuron output of $l$-th layer after activation function $\sigma$. We conduct the Taylor expansion of $\sigma(y)$ function to approximate the output difference with baseline. In other words, we represent the output change $\Delta x_j^{(l+1)} = x_j^{(l+1)} - \tilde x_j^{(l+1)}$ as a Taylor polynomial expansion function of $\Delta y_j^{(l)} = y_j^{(l)} - \tilde y_j^{(l)}$:
\begin{equation}\label{Eq1}
\Delta x_j^{(l+1)}  = \sigma'(y_j^{(l)})\Delta y_j^{(l)} + T^{\gamma},
\end{equation}
where $T^{\gamma}$ represents the high-order terms of $\sigma$ function.
As $\Delta y_j^{(l)} = \sum_{i'} z_{ji}^{(l)}-  \sum_{i'} \tilde z_{ji}^{(l)}$, The Eq. (\ref{Eq1}) is rewritten as,
\begin{equation}\nonumber
 \Delta x_j^{(l+1)}= \sigma'(y_j)(\sum_{i'} z_{ji}^{(l)}-  \sum_{i'} \tilde z_{ji}^{(l)}) +T^{\gamma}.
\end{equation}
 Then the attribution of DeepLIFT of $x_i^{(l)}$ to $x_j^{(l+1)}$ could be reformulated as:
\begin{equation}\nonumber
\begin{aligned}
a_{ij}^{(l)}&= \frac{z_{ji}^{(l)} - \tilde z_{ji}^{(l)}}{\sum_{i'} z_{ji}^{(l)} - \sum_{i'} \tilde z_{ji}^{(l)}}  \Delta x_j^{(l+1)}  \\
& = \frac{z_{ji}^{(l)} - \tilde z_{ji}^{(l)}}{\sum_{i'} z_{ji}^{(l)} - \sum_{i'} \tilde z_{ji}^{(l)}}(\sigma'(y_j)(\sum_{i'} z_{ji}^{(l)}-  \sum_{i'} \tilde z_{ji}^{(l))}) + T^{\gamma} )\\
& = \sigma'(y_j) (z_{ji}^{(l)} - \tilde z_{ji}^{(l)}) +  \frac{z_{ji}^{(l)} - \tilde z_{ji}^{(l)}}{\sum_{i'} z_{ji}^{(l)} - \sum_{i'} \tilde z_{ji}^{(l)}}T^{\gamma} \\
& = T_i^{\alpha} + \frac{z_{ji}^{(l)} - \tilde z_{ji}^{(l)}}{\sum_{i'} z_{ji}^{(l)} - \sum_{i'} \tilde z_{ji}^{(l)}}T^{\gamma}.
\end{aligned}
\end{equation}
Then the conclusion holds.
\end{proof}

\noindent
\textbf{Derivation of $\epsilon$-LRP}\\
First, we give a brief introduction to $\epsilon$-LRP. In $\epsilon$-LRP, $x_i^{(l)}$, $x_j^{(l+1)}$ denote the neuron $i$ at $l$-th layer and the neuron $j$ at $(l+1)$-th layer, respectively, and $x_j^{(l+1)} = \sigma(\sum_i w_{ji}x_i^{(l)} + b_{j})$. Here $w_{ji}$ is the weight parameter, $b_{j}$ is the additive bias, and $\sigma$ is a non-linear activation function. $\epsilon$-LRP propagates the relevance score from $x_j^{(l+1)}$ to $x_i^{(l)}$ as follows:
\begin{equation}\nonumber
a_{ij}^{(l)} = \frac{z_{ji}^{(l)}}{(\sum_{i'} z_{ji'}^{(l)} + b_j + \epsilon sign(\sum_{i'} z_{ji'}^{(l)} + b_j) )} a_{j}^{(l+1)}.
    \end{equation}
Here $z^{(l)}_{ji}$ = $w_{ji}x_i^{(l)}$ is the weighted impact of  $x_i^{(l)}$ to $x_j^{(l+1)}$, and the small quantity $\epsilon$ is added
to the denominator to avoid numerical instabilities. $a^{(l)}_i$ = $\sum_j a^{(l)}_{ij}$ denotes the total relevance score of neuron $x^{(l)}_i$ to all neurons in $(l+1)$-th layer. Next we will use the Taylor attribution framework to reformulate $\epsilon$-LRP.

\begin{proof}
The derivation is similar to DeepLIFT. $x_j^{(l+1)} = \sigma(y_j^{(l)})$ (here $y_j^{(l)} = \sum_{i'} z_{ji}^{(l)}+b_j$) is the $j$-th neuron output of $l$-th layer after activation function $\sigma$. We conduct the Taylor expansion of $\sigma(y)$ function to approximate the output difference w.r.t. baseline. Here, baseline is set to $\bm{\tilde x} = \bm{0}$. Correspondingly, $\tilde y^{(l)}_j = b_j$.  Hence, we represent the output change $\Delta x_j^{(l+1)} = x_j^{(l+1)} - \sigma (b_j)$ as a Taylor polynomial expansion function of $\Delta y_j^{(l)} = \sum_{i'} z_{ji'}^{(l)}$. That is
\begin{equation}\nonumber
\begin{aligned}
\Delta x_j^{(l+1)} & =  \sigma'(y_j) \Delta y_j^{(l)} + T^{\gamma} \\
& = \sigma'(y_j^{(l)}) (\sum_{i'} z_{ji'}^{(l)}) + T^{\gamma},
\end{aligned}
\end{equation}
where $T^{\gamma}$ represents the high-order terms of $\sigma$ function.  Then the attribution of $\epsilon$-LRP of $x_i^{(l)}$ to $x_j^{(l+1)}$ could be reformulated as (we ignore the small quantity $\epsilon$ as it would not effect the attribution in normal numerical case):
\begin{equation}\nonumber
\begin{aligned}
a_{ij}^{(l)} &=  \frac{z_{ji}^{(l)}}{(\sum_{i'} z_{ji'}^{(l)} + b_j)} \Delta x_j^{(l+1)}\\
& = \frac{z_{ji}^{(l)}}{(\sum_{i'} z_{ji'}^{(l)} + b_j)} (\sigma'(y_j^{(l)})\sum_{i'} z_{ji'}^{(l)} + T^{\gamma}).
\end{aligned}
\end{equation}
We can see that, $\epsilon$-LRP could involve the first-order Taylor expansion term (when not considering bias $b_j$). $\epsilon$-LRP considers the overall high-order impact $T^{\gamma}$, which fails to distinguish the high-order terms from each neuron.
\end{proof}

\noindent
\textbf{Proof of Theorem 5}
\begin{proof}
 Integrated Gradients integrates the gradient over the straight line path  from the selected baseline $\bm{\tilde x}$ to input $\bm{x}$. $a_i$ could be formulated as
\begin{equation}\label{Integrated Gradients}\nonumber
a_{i} = ( x_i - \tilde x_i) \int_0^1 \frac{\partial {f(\bm{\tilde x} + \alpha( \bm{x} - \bm{\tilde x}))}}{\partial {x_i}} d\alpha,
\end{equation}
where $ \sum_i a_i = f(\bm{x}) - f(\bm{\tilde x})$. To fit the predefined Taylor expansion form, we rewrite the integral as starting from input $\bm{x}$ to baseline $\bm{\tilde x}$,
\begin{equation}\nonumber
\begin{aligned}
a_i & = (x_i - \tilde x_i) \int_0^1 \frac{\partial {f(\bm{x} + \alpha( \bm{\tilde x} - \bm{x}))}}{\partial {x_i}} d\alpha \\
& = -(\tilde x_i -  x_i ) \int_0^1 \frac{\partial {f(\bm{x} + \alpha( \bm{\tilde x} - \bm{x}))}}{\partial {x_i}} d\alpha \\
& = -\Delta_i \int_0^1 \frac{\partial {f(\bm{x} + \alpha \Delta)}}{\partial {x_i}} d\alpha.
\end{aligned}
\end{equation}
Note that $\int_0^1 \frac{\partial {f(\bm{x} + \alpha \Delta)}}{\partial {x_i}} d\alpha$ is a function of $\Delta$, denoted as $g(\Delta)$. We prove the theorem by conducting Taylor expansion on function $g(\Delta)$ at $\bm{0}$ point. Firstly, the value of $g(\Delta)$ at $\bm{0}$ point is:
\begin{equation}\nonumber
g(\bm{0})  = \int_0^1 \frac{\partial {f(\bm{x})}}{\partial {x_i}} d\alpha\\  =  \frac{\partial {f(\bm{x})}}{\partial {x_i}}.
\end{equation}
Then the first order partial derivative of $g(\Delta)$ with respect to $\Delta_i$ at $\bm{0}$ point is calculated as:
\begin{equation}\nonumber
\begin{aligned}
g_{\Delta_i}(\bm{0}) & = \frac{\partial{\int_0^1 \frac{\partial {f(\bm{x} + \alpha \Delta)}}{\partial {x_i}}d\alpha}}{\partial{\Delta_i}}|_{\Delta = \bm{0}}\\
& =  \int_0^1 \frac{\partial {f(\bm{x} + \alpha \Delta )}}{\partial {x_i}\partial {x_i}}|_{\Delta = \bm{0}}\ \alpha d\alpha\\
& =  \frac{\partial {f(\bm{x})}}{\partial {x_i}\partial {x_i}} \int_0^1 \alpha  d\alpha\\
& = \frac{1}{2}\frac{\partial {f(\bm{x})}}{\partial {x_i}\partial {x_i}}.
\end{aligned}
\end{equation}
Similarly, the first order partial derivative with respect to $\Delta_j$ at $\bm{0}$ point is:
\begin{equation}\nonumber
g_{\Delta_j}(\bm{0}) = \frac{1}{2}\frac{\partial {f(\bm{x})}}{\partial {x_i}\partial {x_j}}.
\end{equation}
So the first order Taylor expansion of $g$ at $\bm{0}$ becomes,
\begin{equation}\nonumber
\begin{aligned}
g(\Delta)& =  g(\bm{0}) + \sum_{k} g_{\Delta_k}(\bm{0}) \Delta_k + \epsilon \\
& =  \frac{\partial {f(\bm{x})}}{\partial {x_i}} + \sum_{k} \frac{1}{2}\frac{\partial {f(\bm{x})}}{\partial {x_i}\partial {x_k}}  \Delta_k + \epsilon,
\end{aligned}
\end{equation}
where $\epsilon$ denotes Taylor expansion error. Correspondingly, as $a_i = - \Delta_i g(\Delta)$, we further obtain (assume $f_{x_ix_j} = f_{x_jx_i}$) ,
\begin{equation}\nonumber
\begin{aligned}
a_i &= -(\frac{\partial {f(\bm{x})}}{\partial {x_i}}\Delta_i + \frac{1}{2}\frac{\partial {f(\bm{x})}}{\partial {x_i}\partial{x_i}} \Delta_i^2 +  \frac{1}{2} \sum_{j\neq i} \frac{\partial {f(\bm{x})}}{\partial {x_i}\partial{x_j}} \Delta_i\Delta_j  + \epsilon)\\
& = -(T_i^{\alpha} + T_i^{\beta_d} + \sum_{\{x_i,x_j\}} \frac{1}{2}T_{\{x_i,x_j\}}^{\beta_t}+ \epsilon).
\end{aligned}
\end{equation}
Hence, in second-order setting, Integrated Gradients assigns first-order term $T_i^{\alpha}$, second-order independent term $T_i^{\beta_d}$ and half of all second-order interactive terms associating with feature subset $\{x_i,x_j\}$, i.e., $\sum_{\{x_i,x_j\}} \frac{1}{2}T_{\{x_i,x_j\}}^{\beta_t}$, to $x_i$.

This could be easily extended to higher order cases. On one hand, the $k$-order derivative with respect to $\Delta_1 ^{k_1}\dots \Delta_i ^{k_i}\dots \Delta_n^{k_n}$ at $\bm{0}$ point is
\begin{equation}\nonumber
\begin{aligned}
g_{\Delta_1 ^{k_1} \dots \Delta_i ^{k_i}\dots \Delta_n^{k_n}}(\bm{0}) & = \frac{\partial{\int_0^1 \frac{\partial {f(\bm{x} + \alpha \Delta)}}{\partial {x_i}}d\alpha}}{\partial {x_i^{k_1}} \dots \partial {x_i^{{k_i}}} \dots \partial {x_n^{k_n}}}|_{\Delta = 0} \\
& =  \int_0^1 \frac{\partial {f(\bm{x} + \alpha \Delta )}}{\partial {x_i^{k_1}} \dots \partial {x_i^{{k_i+1}}} \dots \partial {x_n^{k_n}}}|_{\Delta  = \bm{0}} \ \alpha^k d\alpha \\
& =  \frac{\partial {f(\bm{x})}}{\partial {x_i^{k_1}} \dots \partial {x_i^{{k_i+1}}} \dots \partial {x_n^{k_n}}} \int_0^1 \alpha^k d\alpha \\
&  = \frac{1}{k+1}\frac{\partial {f(\bm{x})}}{\partial {x_i^{k_1}} \dots \partial {x_i^{{k_i+1}}} \dots \partial {x_n^{k_n}}},
\end{aligned}
\end{equation}
where the vector $\bm{k} = [k_1, \dots, k_n]$ satisfies that $\sum_{m} k_m = k$. The coefficient of $\Delta_1 ^{k_1}\dots \Delta_i ^{k_i}\dots \Delta_n^{k_n}$ term is $\frac{1}{k!}\tbinom{k}{k_1,\dots,k_n}g_{\Delta_1 ^{k_1} \dots \Delta_i ^{k_i}\dots \Delta_n^{k_n}}(\bm{0})$, therefore the corresponding Taylor expansion term is,
\begin{equation}\nonumber
 \frac{1}{(k+1)!}\tbinom{k}{k_1,\dots,k_n}\frac{\partial {f(\bm{x})}}{\partial {x_i^{k_1}} \dots \partial {x_i^{{k_i+1}}} \dots \partial {x_n^{k_n}}}\Delta_1 ^{k_1}\dots \Delta_i ^{k_i}\dots \Delta_n^{k_n}.
\end{equation}
As $a_i = -\Delta_i g(\Delta)$, then the corresponding term in $a_i$ is,
\begin{equation}\label{Eq2}
\frac{-1}{(k+1)!}\tbinom{k}{k_1,\dots,k_n} \frac{\partial {f(\bm{x})}}{\partial {x_i^{k_1}} \dots \partial {x_i^{{k_i+1}}} \dots \partial {x_n^{k_n}}}  \Delta_1 ^{k_1}\dots \Delta_i ^{k_i+1}\dots \Delta_n^{k_n}.
\end{equation}
We compare the term in Eq. (\ref{Eq2}) with the coefficient of
$k+1$ order term $\Delta_1 ^{k_1}\dots \Delta_i ^{k_i+1}\dots \Delta_n^{k_n}$ of $f(\bm{\tilde x)} - f(\bm{x})$ expanded at $\bm{x}$, i.e.,
$\frac{1}{(k+1)!}\tbinom{k+1}{k_1,\dots,k_i+1, \dots, k_n} \frac{\partial {f(\bm{x})}}{\partial {x_i^{k_1}} \dots \partial {x_i^{{k_i+1}}} \dots \partial {x_n^{k_n}}}$. We can obtain that when $k_i \neq 0$,
\begin{equation}\nonumber
\frac{\tbinom{k}{k_1,\dots,k_i,\dots, k_n}}{\tbinom{k+1}{k_1,\dots,k_{i+1},\dots, k_n}} =
\frac{k!(k_i+1)!}{(k+1)!k_i!}=  \frac{k_i+1}{k+1}
\end{equation}
That is, Integrated Gradients allocates $\frac{k_i+1}{k+1}$ proportion of the $k+1$ order interactive term $\Delta_1 ^{k_1}\dots \Delta_i ^{k_i+1}\dots \Delta_n^{k_n}$ in Taylor expansion of $f(\bm{x})$ to feature $x_i$. When $k_i = k$, the aforementioned term becomes $k+1$ order independent term, and then corresponding proportion $\frac{k_i+1}{k+1} = 1$.
\end{proof}

\noindent
\textbf{Corollary on Expected Gradients}\\
According to Theorem 5, we can obtain that the attribution $a_i^{IG}(\Delta)$ of Integrated Gradients w.r.t baseline point $\bm{\tilde x}$ can be reformulated as,
\begin{equation}\nonumber
a_i^{IG}(\Delta) = T^{\alpha}_i(\Delta) + T^{\gamma_d}_i(\Delta) + a_i^{\gamma_t}(\Delta).
\end{equation}
The attribution $a_i^{EG}$ of feature $x_i$ in Expected Gradients is an integral of attribution $a_i^{IG}(\Delta)$ over baseline distribution $p_D(\bm{\tilde x})$. We represent the distribution of baseline $p_D(\bm{\tilde x})$ as a distribution of $\Delta$, $p_D(\Delta)$(as $\Delta = \bm{\tilde x} - \bm{x}$). Then the attribution of Expected Gradients $a^{EG}_i$ could be rewritten as
\begin{equation} \nonumber
\begin{aligned}
a^{EG}_i &= \int_{\bm{\tilde x}} a^{IG}_i(\Delta) p_D(\bm{\tilde x}) d\bm{\tilde x}\\
&= \int_{\Delta} a^{IG}_i(\Delta) p_D(\Delta) d\Delta\\
& = \int_{\Delta} T^{\alpha}_i(\Delta) + T^{\gamma_d}_i(\Delta) + a_i^{\gamma_t}(\Delta) d\Delta.
\end{aligned}
\end{equation}
Hence, Expected Gradients averages the first-order term, high-order independent and high-order interactive terms of multiple $\Delta$. This could avoid the case in which Integrated Gradients is dominated by a specific $\Delta$. \\

\section{Appendix B: More experimental results}
\begin{figure*}[t]\centering
\includegraphics[scale=0.5]{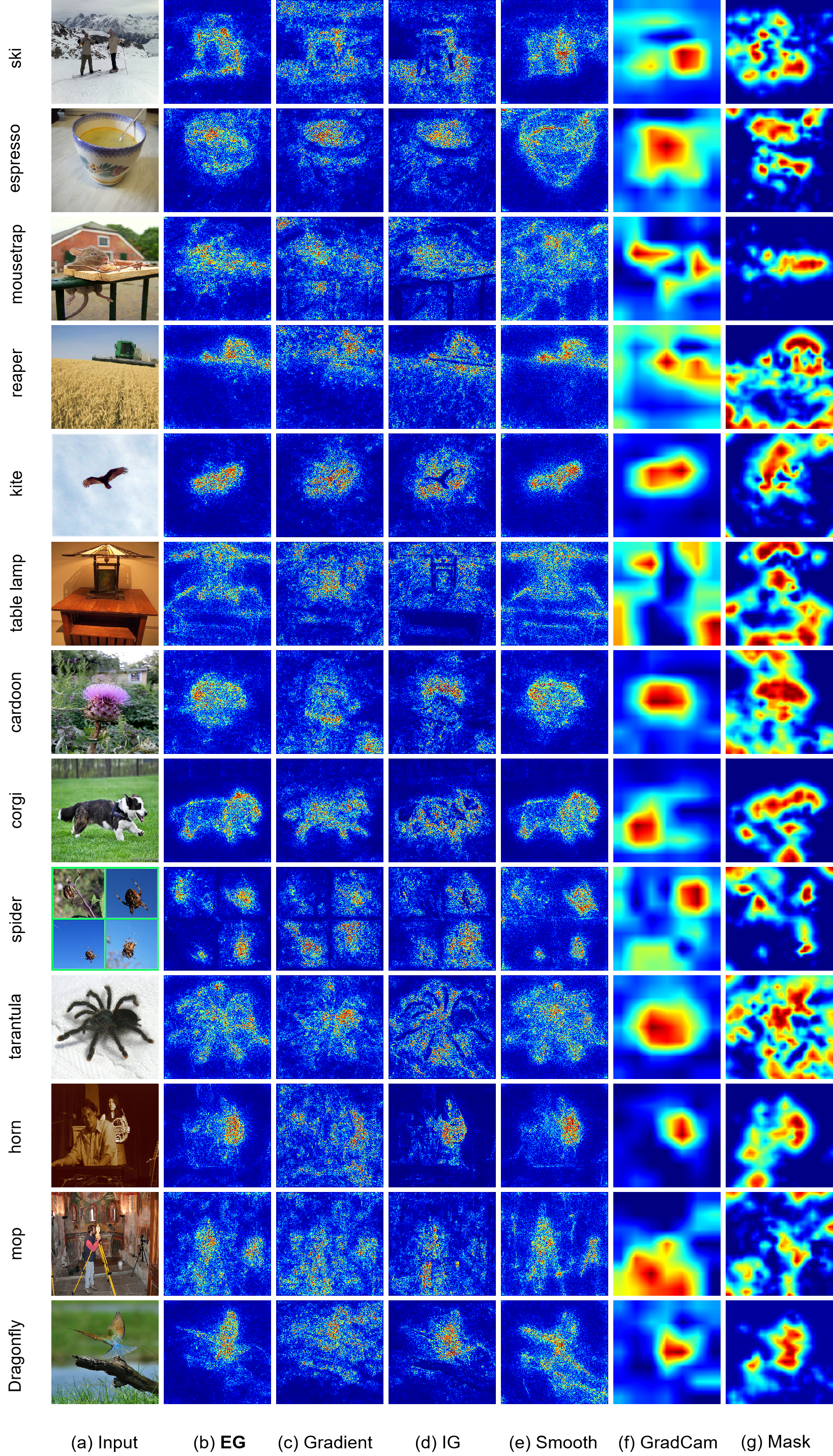}
\caption{Visualization saliency map comparison compared to other attribution methods. }
\label{Visualization saliency map comparison compared to Five state-of-the-art methods}
\end{figure*}

We also qualitatively compare the saliency maps produced using EG with those produced by five state-of-the-art methods, including Gradient, IG, Smooth Grads, Grad-CAM, and Mask learning, see Figure \ref{Visualization saliency map comparison compared to Five state-of-the-art methods}. As shown in Figure \ref{Visualization saliency map comparison compared to Five state-of-the-art methods}, EG has similar saliency maps to Smooth Gradients method in most cases, and has a superior performance in some cases. This interesting fact indicates that Smooth Grads method may have similar theoretical meaning to EG, which provides a new insight to the rationale of Smooth Grads.

\end{document}